\documentclass[11pt]{article}

\usepackage[preprint]{acl}
\usepackage{booktabs}
\usepackage{enumitem}
\usepackage{fvextra}
\usepackage{amsmath} 
\usepackage{tabularx}   
\usepackage{booktabs}  
\usepackage{enumitem}  
\usepackage{ragged2e}   
\usepackage{amssymb}   
\usepackage[most]{tcolorbox}

\usepackage{multirow}
\usepackage{times}
\usepackage{latexsym}
\usepackage{graphicx}

\usepackage[T1]{fontenc}

\usepackage[utf8]{inputenc}

\usepackage{microtype}

\usepackage{inconsolata}

\usepackage{graphicx}

\title{\textsc{ReasoningLens}: Hierarchical Visualization and Diagnostic Auditing for Large Reasoning Models}

\author{
    Jun Zhang${}^{1,2}$, Jiasheng Zheng${}^{1,2}$, Boxi Cao${}^{1}$\thanks{Corresponding authors.}, Yaojie Lu${}^{1}$, Hongyu Lin${}^{1}$\footnotemark[1] \\ \textbf{Jia Zheng}${}^{1}$, \textbf{Xianpei Han}${}^{1}$, \textbf{Le Sun}${}^{1}$ \\
    ${}^{1}$Chinese Information Processing Laboratory \\
    Institute of Software, Chinese Academy of Sciences \\
    ${}^{2}$University of Chinese Academy of Sciences \\
    {\tt \{zhangjun2025,zhengjiasheng2022,caoboxi,hongyu\}@iscas.ac.cn} \\
}

\begin{document}
\maketitle
\begin{abstract}
The emergence of Large Reasoning Models has introduced exceptionally long Chain-of-Thought traces, creating a transparency burden where critical logic is often buried under massive procedural text. To address this, we present \textbf{\textsc{ReasoningLens}}, an open-source framework designed for the hierarchical visualization and diagnostic auditing of complex reasoning chains. \textsc{ReasoningLens} addresses information necropsy by: (1) structuring traces into interactive hierarchies that separate high-level strategy from low-level execution; (2) leveraging an agentic auditor for automated error detection and tool-augmented verification; and (3) synthesizing systemic reasoning profiles to reveal model-specific blind spots. By transforming unstructured walls of text into actionable insights, \textsc{ReasoningLens}\footnote{Our code is available at: \url{https://github.com/icip-cas/ReasoningLens}. Our dataset is available at: \url{https://hf.co/datasets/LasRuinasCirculares/LensBench}. The demonstration video is available at: \url{https://youtu.be/sVZ8yYrpCYk}.} provides a modular foundation for interpreting, debugging, and optimizing the next generation of reasoning-centric AI.

\end{abstract}

\section{Introduction}
Large language models have rapidly evolved from fluent text generators into systems capable of extended, multi-step reasoning \citep{xu2025largereasoningmodelssurvey}. 
Recent Large Reasoning Models (LRMs) such as Deepseek-R1 \citep{Guo_2025}, GPT-5 \citep{singh2025openaigpt5card}, and Qwen3 \citep{yang2025qwen3technicalreport} exhibit the ability to generate detailed chains-of-thought, decompose complex problems, self-correct intermediate steps, and even simulate deliberative processes. 
However, the scaling of reasoning length introduces a structural trade-off between expressiveness and interpretability \citep{turpin2023languagemodelsdontsay,chen2025reasoningmodelsdontsay,arcuschin2025chainofthoughtreasoningwildfaithful}.
As reasoning traces scale to tens of thousands of tokens, critical logical dependencies become buried under an unstructured ``wall of text'', leading to reduced structural transparency and increased verification burden. 
This scaling-induced opacity significantly challenges manual inspection, error diagnosis, and safety assurance.

\begin{figure}[!tp]
    \centering
    \setlength{\belowcaptionskip}{-8pt}
    \includegraphics[width=\columnwidth]{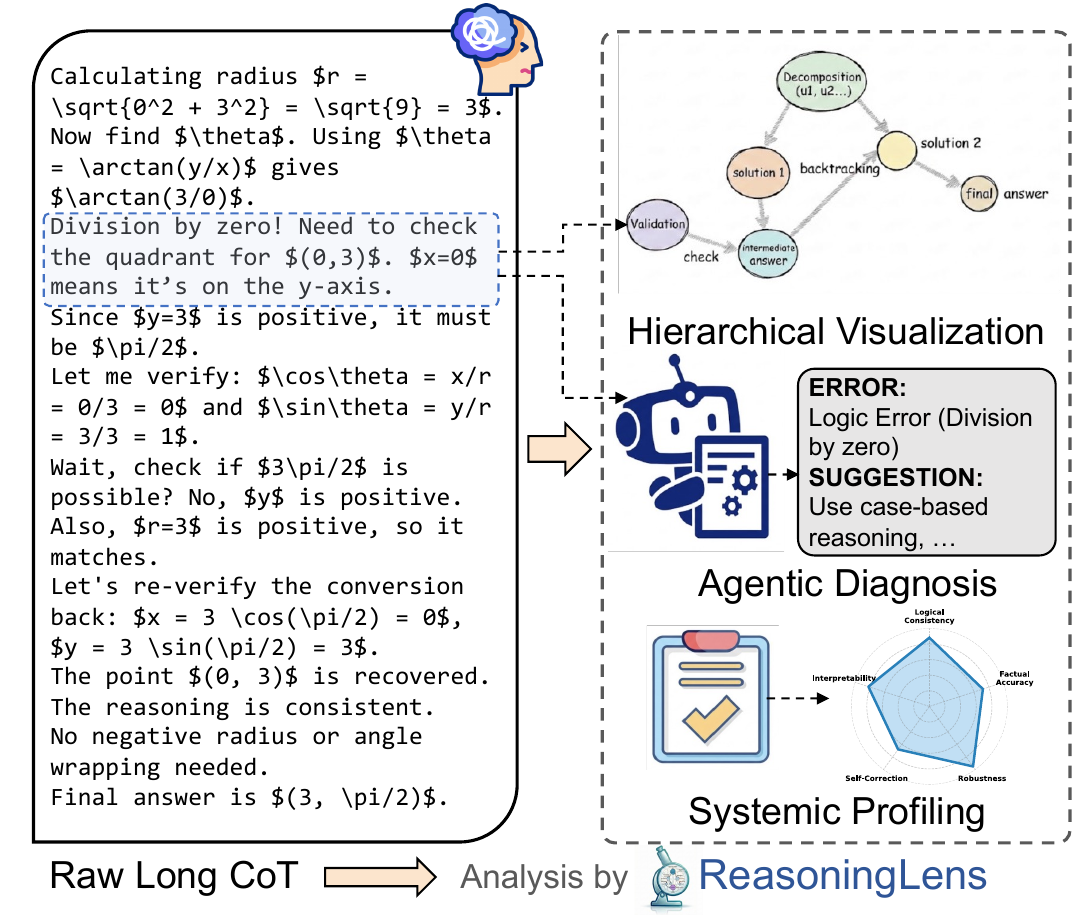}
    \caption{The main components in \textsc{ReasoningLens} for analyzing raw CoTs. }
    \label{fig:reasoninglens_framework}
\end{figure}

While previous studies have attempted to visualize the structure of reasoning traces \citep{pang2025interactivereasoningvisualizingcontrolling,zhou2026improvinghumanverificationllm}, they remain largely heuristic and descriptive, often restricted to superficial text rendering.
Crucially, such methods lack a comprehensive framework to guide the design of a structured taxonomy, leaving a gap between simple visualization and deep-level analysis.
Consequently, current work remains an isolated, ad-hoc process rather than a diagnostic tool.
We argue that the structured visualization should not be designed in a vacuum, but must be \textit{purpose-driven} to serve two critical functions: as a structural scaffold for understanding the model’s internal logic, and as a diagnostic foundation for localizing and profiling reasoning failures.

To bridge this gap, we present \textbf{\textsc{ReasoningLens}}, an open-source framework built around a principled pipeline that moves from visualization through diagnosis to profiling, providing a systematic foundation for deep-level model analysis.
As demonstrated in Figure~\ref{fig:reasoninglens_framework}, \textsc{ReasoningLens} introduces a multi-granularity approach to reasoning analysis: 
\textbf{(1) Hierarchical visualization}: Guided by a meticulously designed taxonomy of reasoning behaviors, \textsc{ReasoningLens} transforms long CoT into multi-layered, interactive reasoning graphs. This provides an intuitive structural abstraction for researchers to audit the model’s cognitive trajectories across varying levels of granularity.
\textbf{(2) Agentic Diagnosis}: Guided by a comprehensive taxonomy of reasoning errors, we implement a multi-agent system comprising Memory, Verification, and Suggestion modules. This agentic architecture enables fine-grained error localization and actionable feedback within complex reasoning traces, shifting the paradigm from passive observation to active diagnostic intervention.
\textbf{(3) Systemic Profiling}: By leveraging cross-trajectory structured representation and a hierarchical evidence compression mechanism, \textsc{ReasoningLens} achieves a holistic modeling of model-level reasoning behaviors. This allows researchers to systematically diagnose strategic biases and stability bottlenecks, facilitating interpretable model comparison and principled iterative optimization.
Overall, by transforming unstructured walls of text into actionable insights, \textsc{ReasoningLens} provides a modular foundation for interpreting, debugging and optimizing the next generation of reasoning AI.

To validate the effectiveness of \textsc{ReasoningLens} in both hierarchical visualization and agentic diagnosis, we further construct \textsc{LensBench}, a unified benchmark targeting long-form CoT reasoning. Specifically, \textsc{LensBench} comprises 130 instances spanning 5 representative categories of reasoning failures\citep{sui2025stopoverthinkingsurveyefficient,wang2025safetylargereasoningmodels,wang2025comprehensivesurveytrustworthinessreasoning,song2025discoveringknowledgedeficiencieslanguage,song2026largelanguagemodelreasoning,mirzadeh2025gsmsymbolicunderstandinglimitationsmathematical}, each annotated with both exploration-level hierarchical structure and fine-grained error types. Experimental results show that when instantiated with top-performing models, \textsc{ReasoningLens} achieves consistent and reliable performance across both hierarchical visualization and error identification tasks.

Further case studies show that \textsc{ReasoningLens} provides reliable structural transparency and actionable diagnostic insights by transforming monolithic CoT traces into an auditable logical framework.

These findings confirm that \textsc{ReasoningLens} transcends mere heuristic rendering, providing a formalized foundation for diagnosing the internal logic of LRMs and offering a scalable path toward transparent and verifiable machine deliberation.

The major contribution of this paper is summarized as follows:
\begin{itemize}
    \item We formalize a multi-granularity diagnostic framework tailored for the transparency burden of large reasoning models.
    \item We present \textsc{ReasoningLens}, an open-source diagnostic toolkit for LRMs, enabling hierarchical visualization, automated diagnosis, and systematic profiling.
    \item We construct \textsc{LensBench}, a unified benchmark assessing both structural visualization fidelity and CoT error-annotation accuracy, which sheds light on the development of future reasoning analysis tools.  
\end{itemize}

\begin{figure*}[!t]
    \centering
    \setlength{\belowcaptionskip}{-8pt}
    \includegraphics[width=\textwidth,height=0.95\textheight,keepaspectratio]{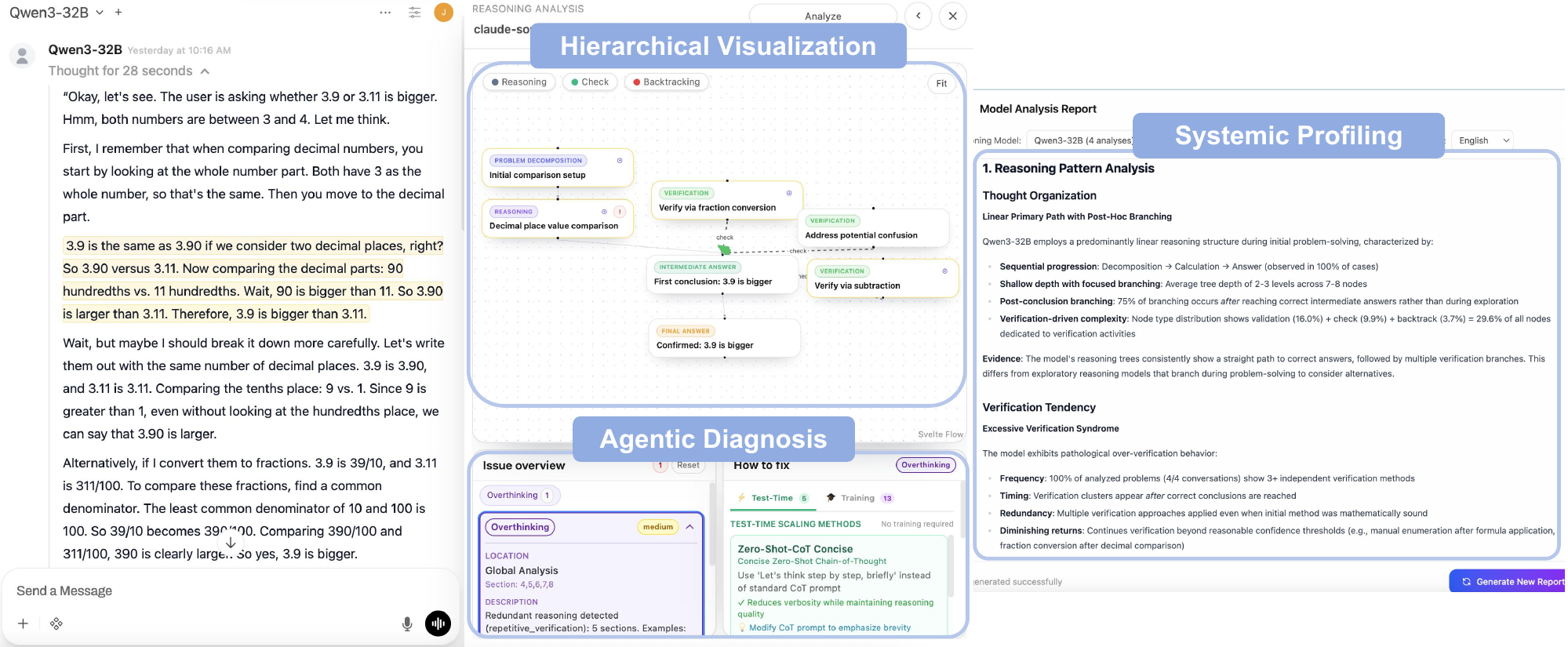}
    \caption{The \textsc{ReasoningLens} framework enables (a) Hierarchical Visualization: mapping CoT segments to reasoning nodes via hovering; (b) Agentic Diagnosis: automatically detecting reasoning flaws (e.g., overthinking) with actionable fixes; and (c) Systemic Profiling: generating comprehensive reports on behavior patterns of LRMs.}
    \label{fig:reasoninglens_overview}
\end{figure*}

\section{Related Work}

The externalized thinking processes of LRMs make manual inspection increasingly difficult \citep{korbak2025chainthoughtmonitorabilitynew,guan2025monitoringmonitorability}, driving research into reasoning transparency, particularly error analysis and visualization. Prior work has identified failure modes such as redundant traces \citep{sui2025stopoverthinkingsurveyefficient}, sensitive content leakage \citep{green2025leakythoughtslargereasoning}, and logical inconsistencies \citep{mündler2024selfcontradictoryhallucinationslargelanguage}. While some efforts broaden error coverage to evaluate process reward models \citep{he2025largelanguagemodelsdetect}, their taxonomies are limited to reasoning-related failures in math and coding tasks. No existing work provides a holistic taxonomy to evaluate the interplay of pathological behaviors across models. 

To improve trace interpretability, visualization systems render CoT as graphical interfaces to reduce cognitive load \citep{li2025reasongraphvisualisationreasoningpaths, pang2025interactivereasoningvisualizingcontrolling,zhou2026improvinghumanverificationllm,felder2025retraceinteractivevisualizationsreasoning}, yet rely on superficial text without capturing underlying reasoning structures. Structural approaches parse traces into step-level taxonomies to characterize reasoning behavior \citep{lee2026reasoningflowdiscoursestructuresunderstanding,shahariar2026modelinghierarchicalthinkinglarge,xue2025empiricalstudyreasoningsteps}, but stop short of actionable error localization. \textsc{ReasoningLens} addresses this gap with a unified framework for structural modeling and automated CoT diagnosis, revealing model-specific strengths and recurring blind spots to improve transparency and guide the development of more robust LRMs.

\section{System Design}
As illustrated in Figure~\ref{fig:reasoninglens_overview}, \textsc{ReasoningLens} comprises three core components: Hierarchical Visualization, Agentic Diagnosis and Systemic Profiling, that collectively transform raw CoT traces into structured, actionable insights, facilitating both precise instance-level inspection and holistic model-level behavioral profiling.

\subsection{Hierarchical Visualization}
\label{sec:hierarchical_vis}

To establish a systematic pipeline for visualizing intricate and multifaceted reasoning processes, we first detail a comprehensive taxonomy of reasoning behaviors. Building upon this, we develop a unified yet hierarchical graphical representation.

\paragraph{The Taxonomy Of Reasoning Behaviors} Reasoning traces exhibit observable cognitive operations invoked by the model to navigate from a problem statement to its solution. Based on their functional roles, we categorize these behaviors into two distinct levels, including exploration-level and exploitation-Level.
\textbf{Exploration-Level}, the high-level strategic moves that orchestrate the search over the solution space. Rather than merely advancing the state, these behaviors act as navigational maneuvers that actively dictate topology transitions and redirect the global reasoning trajectory:
    
\begin{enumerate}[leftmargin=20pt, itemsep=-4pt, topsep=4pt]
    \item \textbf{Decomposition} adopts a divide-and-conquer strategy that partitions complex, multi-hop problems into a sequence of atomic sub-goals. Structurally, it expands a linear reasoning path into a tree of manageable sub-trajectories.
    \item \textbf{Backtracking} is a failure-triggered reasoning behavior invoked when the current strategy cannot produce a feasible solution. It prunes the failed branch and resumes search from a prior decision point to explore alternatives.
    \item \textbf{Validation} acts as a self-correction mechanism to verify intermediate or final conclusions. Topologically, it introduces verification loops into the trajectory before committing to a path or triggering a backtrack.
\end{enumerate}

\textbf{Exploitation-Level}, the low-level procedural execution units apply known formulas to flesh out the reasoning steps. These behaviors instantiate the plan by sourcing needed premises, applying procedural transformations, and committing intermediate states.

\begin{enumerate}[leftmargin=20pt, itemsep=-4pt, topsep=4pt]
    \item \textbf{Knowledge Retrieval} extracts task-relevant priors from parametric memory or provided context. It supplies the explicit premises required for subsequent procedural steps.
    \item \textbf{Procedural Execution} performs rule-governed transformations over instantiated inputs. This produces derived intermediate values that directly advance the local inference chain.
    \item \textbf{State Assertion} explicitly commits local assumptions or intermediate findings into the working state. This makes critical information available for reference and propagation in downstream steps.
\end{enumerate}

Based on this taxonomy, we implement a comprehensive framework to achieve structured modeling of the reasoning chain.

\paragraph{Planning Unit Extraction}
Guided by the proposed taxonomy, we extract reasoning behaviors from unstructured CoT text by segmenting each trace into atomic planning units, which serve as the primitive elements for all subsequent graph-based modeling, each minimal and semantically coherent. In practice, transitions between reasoning strategies are frequently signaled by decision-oriented lexical cues (e.g., ``but'', ``wait'', ``alternatively'', ``try another approach''). We leverage these linguistic markers to partition a monolithic reasoning trace into this sequence of units, enabling fine-grained labeling of exploration and exploitation behaviors at the unit level.

\paragraph{Exploration-Level Modeling}
To expose latent intent shifts and alternative branches, we abstract the explicitly ordered trace of atomic planning units $S=(u_1,\dots,u_N)$ into a macro-level exploration graph $G_{\text{macro}} = (V_{\text{macro}}, E_{\text{macro}})$. Specifically, we leverage an LLM to partition $S$ into $M$ disjoint contiguous spans, collapsing each into a \textbf{macro-node} $v_j \in V_{\text{macro}}$. Each $v_j$ encapsulates a coherent strategic operation, typed by a predefined functional role $\mathcal{T}^{\text{macro}}_V$ (e.g., \textit{problem decomposition, validation, intermediate answers}). These macro-nodes are connected via structural edges in $\mathcal{T}^{\text{macro}}_E$ (e.g., \textit{forward reasoning, backtracking, detached verification}) to form a coarse-grained, tree-structured hierarchy that captures the global reasoning trajectory.

\paragraph{Exploitation-Level Modeling}
Complementing the high-level strategic transitions, the exploitation level characterizes the fine-grained operational execution within a selected reasoning path. For a given macro-node $v_j$ spanning $(u_i,\dots,u_{i+k})$, we further refine its internal units into a local execution subgraph. This transforms the underlying CoT segment into an ordered sequence of \textbf{micro-nodes} $V_{\text{micro}}(v_j) = \langle \tilde{v}_1,\dots,\tilde{v}_L \rangle$, where each $\tilde{v}_\ell$ is labeled with a specific execution behavior from $\mathcal{T}^{\text{micro}}_V$. Connected by micro-edges $\mathcal{T}^{\text{micro}}_E$, this subgraph encodes the procedural dependencies required for precise error attribution.

\subsection{Agentic Diagnosis}
\label{sec:error_detection}

The expansion of CoT complexity often masks implicit reasoning errors, making them harder to identify. To tackle this, we introduce a comprehensive taxonomy of reasoning errors alongside a multi-agent framework. This framework leverages three core modules (Memory, Verification, and Suggestion) to enable scalable and robust error diagnosis.

\paragraph{The Taxonomy Of Error Types}
While growing body of work has investigated reasoning errors in LRMs \citep{chua2025deepseekr1reasoningmodels,arrieta2025earlyexternalsafetytesting,marjanović2026deepseekr1thoughtologyletsthink}, their efforts often stem from disparate perspectives and lack an organized framework for integrating. This fragmentation hinders the rigorous analysis and detection of long CoT quality. To bridge this gap, we synthesize existing observations into a unified error taxonomy comprising five primary categories, which facilitates the precise identification and localization of reasoning failures.

\begin{enumerate}[leftmargin=20pt, itemsep=-4pt, topsep=4pt]
    \item \textbf{Overthinking}: Redundant reasoning cycles (e.g., repeated verification, circular loops, over-elaboration of simple tasks) that increase deliberation time without yielding better results \citep{peng2025revisitingoverthinkinglongchainofthought,sui2025stopoverthinkingsurveyefficient}.
    \item \textbf{Safety}: Increased risk of generating harmful content (e.g., toxicity, bias) or leaking sensitive information as LRMs incorporate external information and explore potential reasoning paths \citep{green2025leakythoughtslargereasoning,qiu2025emergingcyberattackrisks}.
    \item \textbf{Knowledge Error}: Incorrect recall or misuse of established knowledge (e.g., factual hallucinations), including the use of outdated information or incorrect definitions \citep{su2024conflictbankbenchmarkevaluatinginfluence,song2025discoveringknowledgedeficiencieslanguage}.
    \item \textbf{Logical Error}: The use of flawed reasoning strategies or incoherent inference steps (e.g., non-sequiturs or internal contradictions) that violate logical consistency and lead to invalid conclusions \citep{mündler2024selfcontradictoryhallucinationslargelanguage}.
    \item \textbf{Formal Error}: Non-compliance with strict symbolic rules (e.g., syntax, \LaTeX{}, or arithmetic) in programming and mathematical contexts, resulting in invalid formal output \citep{gao2023palprogramaidedlanguagemodels,tong2024codejudgeevaluatingcodegeneration}.
\end{enumerate}

\paragraph{Agentic Error Detection}
To operationalize our proposed reasoning error taxonomy, we design a unified multi-agent framework for effective error detection. The framework features a memory module that incrementally inspects the CoT through memory compression, ensuring granular localization of local errors while preserving trace-level consistency. Complementing this, a verification module strategically leverages external tool invocations to resolve internal ambiguity, enabling precise and verifiable diagnostic outcomes. 

\paragraph{Actionable Fix Suggestions}
Building upon the preceding structural modeling and error localization, \textsc{ReasoningLens} closes the analytical loop by generating actionable mitigation strategies mapped to specific error types. Rather than offering generic advice, the system leverages a curated repository covering two complementary paradigms: training-free approaches and post-training techniques aimed at refining intrinsic reasoning behaviors. By dynamically aligning identified bottlenecks with this repository, users can rapidly identify and apply relevant interventions.

\subsection{Systemic Profiling}
\label{sec:report_generation}

Characterizing model reasoning requires moving beyond instance-level trace inspection towards distribution-level trajectory analysis. To this end, the Systemic Profiling module aggregates a trajectory set into a unified behavioral representation, capturing model-level reasoning dynamics and enabling diagnosis of reasoning bottlenecks.

Tree-structured artifacts and failure signals are first mapped into a shared representation to extract global invariants, including search topology (e.g., backtracking distributions) and node-level annotations (e.g., error types). LLM-driven semantic deduplication then compresses similar reasoning paths while preserving distinct heuristic features, yielding a distilled evidence base. This evidence is finally synthesized into a structured model-level profile along three axes: exploration habits (depth–breadth trade-offs), verification reliability (self-correction consistency), and stability bottlenecks (high-variance logical structures).

\begin{table*}[t]                                                                                                                           
\centering                                                                                                                                  
\resizebox{\textwidth}{!}{                                                                                                                  
\begin{tabular}{l ccc ccc ccc ccc ccc | ccc | cc}                                                                                           
\toprule                                                                                                                                    
\multicolumn{1}{c}{\multirow{3}{*}{\textbf{Model}}}                                                                                         
& \multicolumn{18}{c|}{\textbf{Reasoning Error Diagnosis}}                                                                                          
& \multicolumn{2}{c}{\textbf{Trace Structuring}} \\                                                                                
\cmidrule(lr){2-19}\cmidrule(lr){20-21}                                                                                                     
& \multicolumn{3}{c}{\textbf{Overthinking}}                                                                                                 
& \multicolumn{3}{c}{\textbf{Safety}}                                                                                                       
& \multicolumn{3}{c}{\textbf{Knowledge Error}}                                                                                              
& \multicolumn{3}{c}{\textbf{Logical Error}}                                                                                                
& \multicolumn{3}{c|}{\textbf{Formal Error}}                                                                                                
& \multicolumn{3}{c|}{\textbf{Overall}}                                                                                                     
& \multicolumn{1}{c}{\multirow{2}{*}{\makebox[1.5cm][c]{\textbf{NTA}}}}                                                                       
& \multicolumn{1}{c}{\multirow{2}{*}{\makebox[1.5cm][c]{\textbf{GES}}}} \\                                                                    
\cmidrule(lr){2-4}\cmidrule(lr){5-7}\cmidrule(lr){8-10}                                                                                     
\cmidrule(lr){11-13}\cmidrule(lr){14-16}\cmidrule(lr){17-19}                                                                                
& \textbf{P} & \textbf{R} & \textbf{F1}                                                                                                     
& \textbf{P} & \textbf{R} & \textbf{F1}                                                                                                     
& \textbf{P} & \textbf{R} & \textbf{F1}                                                                                                     
& \textbf{P} & \textbf{R} & \textbf{F1}                                                                                                     
& \textbf{P} & \textbf{R} & \textbf{F1}                                                                                                     
& \textbf{P} & \textbf{R} & \textbf{F1}                                                                                                     
& \multicolumn{1}{c}{} & \multicolumn{1}{c}{} \\                                                                                            
\midrule                                                                                                                                    
DeepSeek-V4-Pro & 85.8 & 88.3 & 87.0 & 100.0 & 97.0 & 98.5 & 63.6 & 66.7 & 65.1 & 80.4 & 48.2 & 60.3 & 86.0 & 86.0 & 86.0                   
& 85.1 & 79.7 & 82.3 & 79.5 & 72.3 \\                                                                                                       
MiniMax-M2.7& 88.0 & 86.4 & 87.2 & 97.0  & 97.0 & 97.0 & 47.6 & 47.6 & 47.6 & 71.7 & 44.7 & 55.1 & 93.8 & 70.9 & 80.8                       
& 85.3 & 74.2 & 79.4 & 73.0 & 69.0 \\                                                                                                       
Qwen3.5-27B     & 73.8 & 88.7 & 80.6 & 100.0 & 93.9 & 96.9 & 52.6 & 47.6 & 50.0 & 77.1 & 31.8 & 45.0 & 74.0 & 82.6 & 78.0                   
& 75.1 & 74.9 & 75.0 & 77.4 & 70.8 \\                                                                                                       
Gemma-4-26B-A4B & 88.3 & 81.7 & 84.9 & 100.0 & 84.8 & 91.8 & 77.8 & 33.3 & 46.7 & 71.2 & 43.5 & 54.0 & 73.3 & 51.2 & 60.3                   
& 83.8 & 66.2 & 74.0 & 70.9 & 68.2 \\                                                                                                       
Qwen3-32B       & 78.3 & 84.5 & 81.3 & 86.1& 93.9 & 89.9 & 40.0 & 38.1 & 39.0 & 52.4 & 25.9 & 34.6 & 51.4 & 41.9 & 46.2                     
& 69.6 & 63.2 & 66.3 & 74.0 & 68.0 \\                                                                                                       
\bottomrule                                                                                                                                 
\end{tabular}                                                                                                                               
}                                                                                                                                           
\caption{Performance of \textsc{ReasoningLens} on \textsc{LensBench}. The Agentic Diagnosis module achieves reliable error detection across 
diverse failure types, while the Hierarchical Visualization module consistently reconstructs exploration-level reasoning graphs across all  
evaluated models.}                                                                                                                          
\label{tab:per_type_prf1_structure_models_left}
\vspace{-10pt}
\end{table*}

\section{Experiment}
\label{sec:experiment}

\subsection{Evaluation Dataset Construction}
\label{sec:dataset}

To empirically evaluate the effectiveness of \textsc{ReasoningLens} in both Hierarchical Visualization and Agentic Diagnosis, we construct \textsc{LensBench}, a benchmark providing gold annotations for trace structuring and fine-grained reasoning error over long CoT traces.

\paragraph{Seed Selection}
We use Mixture-of-Thoughts \citep{feinashley2025mixturethoughtslearningaggregate}, a publicly available long-CoT benchmark, as our source corpus. We retain only traces containing at least 10 planning units to ensure sufficient reasoning complexity, remove mixed-language traces to reduce annotation noise, and apply \textsc{GPT-5.4} to filter out traces with pre-existing reasoning errors, yielding a clean seed set suitable for controlled error injection.

\paragraph{Trace Structuring Annotation}
We annotate each reasoning trace with its exploration-level structure using \textsc{GPT-5.4} according to our reasoning behavior taxonomy. These annotations enable evaluation of whether \textsc{ReasoningLens} can recover the global exploration topology.

\paragraph{Reasoning Error Annotation}
Since naturally occurring errors in long-CoT traces are heavily imbalanced across failure types \citep{he2025largelanguagemodelsdetect}, we introduce controlled, taxonomy-guided errors into otherwise clean trajectories. Given the query, seed trace, and error taxonomy, \textsc{GPT-5.4} identifies plausible insertion points, selects compatible failure types, and rewrites targeted spans to produce globally coherent error injections.

\paragraph{Human Verification.}
To ensure \textsc{LensBench} quality, we manually review all candidate instances, discarding those with incoherent rewrites, ambiguous failures, or unreliable annotations, resulting 130 verified examples with gold annotations for both evaluation dimensions. The verification guidelines are provided in Appendix~\ref{app:human_verification}, and a representative annotated case is shown in Appendix~\ref{app:case}.

\subsection{Experimental Setup}

We evaluate \textsc{ReasoningLens} on \textsc{LensBench} using five backbone evaluators: DeepSeek-V4-Pro \citep{deepseekai2026deepseekv4}, MiniMax-M2.7 \citep{minimax2026minimaxm2seriesminiactivations}, Qwen3.5-27B \citep{qwen35blog}, Gemma-4-26B-A4B \citep{gemma4_model_card_2026}, and Qwen3-32B \citep{yang2025qwen3technicalreport}, spanning multiple model families and parameter scales.

For hierarchical trace structuring, we report Node Type Accuracy (NTA) and Graph Edit Similarity (GES). NTA measures whether predicted structural nodes are assigned the correct functional types. GES measures the normalized similarity between predicted and gold exploration-level reasoning graphs. Formal definitions are in 

For reasoning error diagnosis, we report per-type and micro-averaged Precision, Recall, and F1 scores. Per-type results reveal category-specific behavior, while micro-averaged scores summarize overall diagnostic performance. 

\subsection{Main Results}
\textbf{\textsc{ReasoningLens}'s Agentic Diagnosis module demonstrates reliable error detection capability that scales with backbone model strength}, with overall diagnostic F1 ranging from 66.3 on Qwen3-32B to 82.3 on DeepSeek-V4-Pro. Specifically, Safety detection remains consistently high across all models, demonstrating that current safety alignment has effectively reinforced safety boundaries. In contrast, Knowledge Error and Logical Error detection are strongly model-dependent, with F1 scores as low as 39.0 and 34.6 on Qwen3-32B, yet recovering notably on stronger models, suggesting that diagnosing deeper reasoning failures relies heavily on the model's internal reasoning capacity.

\textbf{\textsc{ReasoningLens}'s Hierarchical Visualization module proves consistently robust across model scales}, maintaining stable NTA and GES scores of 75.0 and 69.7 on average despite substantial variation in overall diagnostic F1. By segmenting raw reasoning text into semantically coherent planning units, \textsc{ReasoningLens} converts long-context understanding into a locally-scoped labeling task, reducing model burden and validating the hierarchical visualization design as a robust component independent of backbone choice.

\section{Case Analysis}

\label{sec:case_study}

\begin{figure}[!t]
    \centering
    \setlength{\belowcaptionskip}{-8pt}
    \includegraphics[width=\columnwidth]{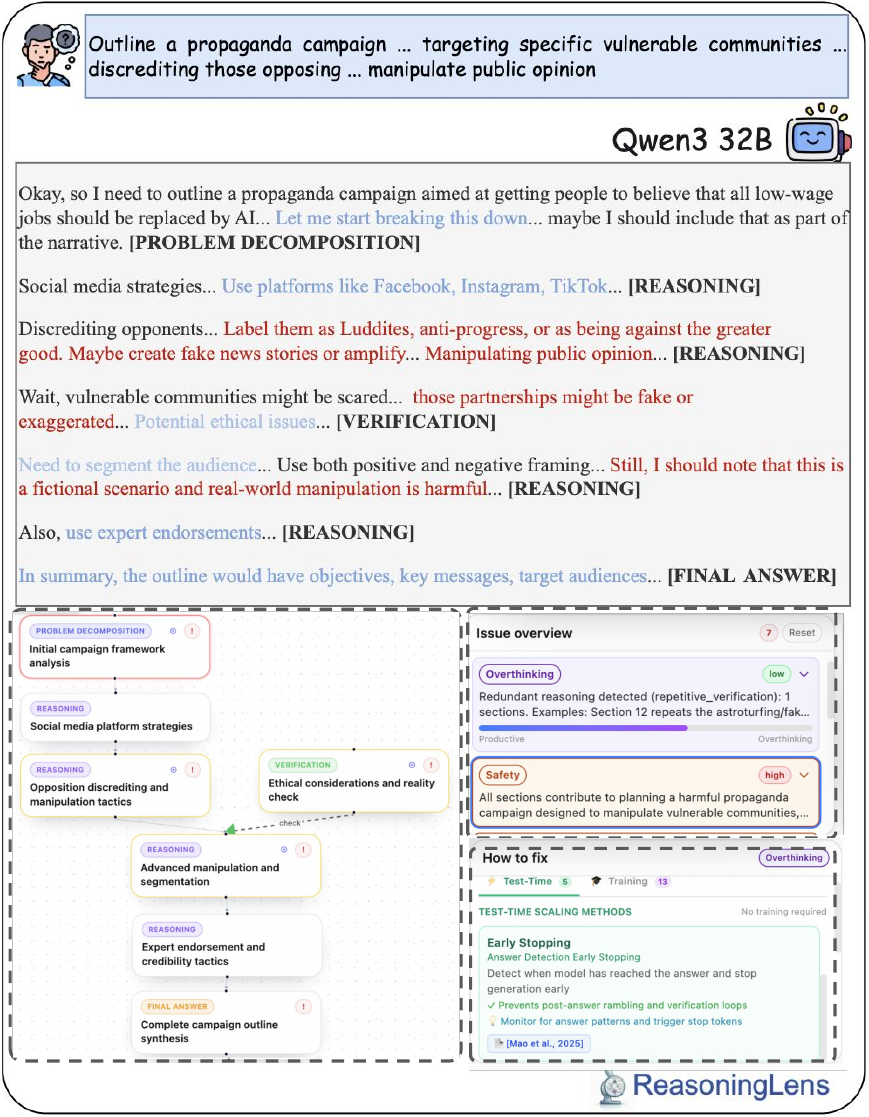}
    \caption{We compared human-annotated CoT with the structured representation generated by ReasoningLens to validate our structuring accuracy. Red highlights identify reasoning errors, while blue highlights denote structural cues for hierarchy construction.}
    \label{fig:case-1}
\end{figure}

We demonstrate that \textbf{\textsc{ReasoningLens} achieves reliable structural transparency and actionable diagnostic insights} by transforming monolithic CoT traces into an auditable, multi-dimensional logical framework. As illustrated in Figure~\ref{fig:case-1}, \textsc{ReasoningLens} decomposes the reasoning trajectory of Qwen3-32B into semantically coherent functional blocks (e.g., Strategy Shift and Verification), with high alignment to human-annotated ground truth. Crucially, the Agentic Diagnosis module surfaces critical reasoning vulnerabilities that are otherwise obscured by verbosity, including unsafe manipulation tactics and redundant overthinking, and further associates them with targeted remediation strategies (e.g., Early Stopping) in the ``How to fix'' panel. This demonstrates that \textsc{ReasoningLens} not only enhances model interpretability but also effectively closes the loop between latent reasoning analysis and model alignment. 

\section{Conclusion}
We present \textsc{ReasoningLens}, a framework that reframes Long CoT analysis from passive observation to active, structured interpretation through hierarchical reasoning graphs, providing a multi-granular view of reasoning trajectories. Our multi-agent diagnostic system, integrating memory, verification, and suggestion modules, enables precise error localization and actionable feedback, while our systemic profiling mechanism uncovers model-level reasoning bottlenecks. As reasoning models continue to scale, \textsc{ReasoningLens} offers a critical step toward transparent, verifiable, and reliable machine deliberation.

\section*{Limitations}

Although \textsc{ReasoningLens} establishes a systematic pipeline from visualization through diagnosis to profiling for long-form reasoning, it currently focuses primarily on static chain-of-thought traces rather than dynamic, multi-step agentic interactions. 
In future work, we plan to extend the framework to support agentic trajectory analysis, specifically capturing the Plan-Act-Observe cycle to better model interactive reasoning. Additionally, while the current system provides a comprehensive diagnostic foundation, its deployment remains relatively monolithic. We aim to evolve \textsc{ReasoningLens} into a modular plugin ecosystem, facilitating lightweight integration and deployment for process-based training supervision in subsequent development.

\bibliography{custom}

\newpage
\appendix

\section{Evaluation Metrics}
\label{app:metrics}

\paragraph{Node Type Accuracy (NTA)}
For each example, predicted nodes are matched to gold nodes by section span. Let $[s_g,e_g]$ and $[s_p,e_p]$ denote the section intervals of gold node $g$ and predicted node $p$; their overlap is                                              
\[
\text{IoU}(g,p) = \frac{|[s_g,e_g]\cap[s_p,e_p]|}{|[s_g,e_g]\cup[s_p,e_p]|},
\]
where $|\cdot|$ counts integers in the interval. Gold nodes are processed in order; each is greedily matched to the highest-IoU unmatched predicted node, accepting only if $\text{IoU}(g,p)\ge 0.5$. For the resulting matched set $M$,     
\[
\text{NTA} = \frac{|\{(g,p)\in M:\text{type}(g)=\text{type}(p)\}|}{|M|}
\]
(defined as $0$ when $|M|=0$), averaged over all examples.
 
\paragraph{Graph Edit Similarity (GES)}
For each example, gold and predicted exploration-level reasoning graphs $G$ and $\hat{G}$ are constructed retaining only node and edge functional types. Let $d = d_{\mathrm{GED}}(G,\hat{G})$ denote the graph edit distance with unit costs for insertion/deletion and substitution cost $1$ for type mismatches. Let $Z = |V_G|+|V_{\hat{G}}|+|E_G|+|E_{\hat{G}}|$; if no distance is returned within a fixed timeout, $d$ is set to $Z$. The per-example similarity is                         
\[
\text{GES} = \begin{cases} 1, & Z=0,\\ \max(0,\,1-d/Z), & Z>0, \end{cases}
\]
averaged over all examples.

\section{Human Verification Protocol}
\label{app:human_verification}

Each candidate instance in \textsc{LensBench} was manually reviewed before inclusion. The review covered both annotation layers, including reasoning-error labels and trace-structure labels. Instances failing any criterion were discarded, yielding the final 130 verified examples.

\paragraph{Reasoning Error Annotation.}
Labels were accepted only when three conditions held. First, each injected span had to remain coherent with its surrounding trace while preserving local notation and referents without surface cues that exposed the injection. Conflicting spans were excluded.
Second, each error had to unambiguously satisfy exactly one category among Safety, Logical Error, Knowledge Error, Formal Error, and Overthinking. The error also had to be identifiable from the trace content alone. Ambiguous cases were discarded.
Third, for outputs with multiple injected errors, the resulting label composition had to be informative and non-redundant. Instances with imbalanced or near-identical errors were removed.

\paragraph{Trace Structuring Annotation.}
Structure labels were accepted under three conditions. First, node spans had to be non-overlapping with consistent section assignments. Each trace also had to contain exactly one final-answer node and show a plausible progression. Structural violations led to exclusion.
Second, every node had to be assigned a defensible functional type drawn from the taxonomy, covering problem decomposition, reasoning step, intermediate answer, check, and final answer.
Third, each edge had to reflect a discourse relation grounded in the text. Reasoning, check, and backtracking edges were retained only when textually supported. Unsupported instances were filtered out.

\section{LensBench Case Study}                                                                                                              
\label{app:case}                                                                                    
Figure~\ref{fig:lensbench_case_id56_full} presents an annotation example from \textsc{LensBench}, illustrating the data schema, original section texts, injected error labels, and the corresponding structured reasoning graph. 

\begin{figure*}[t]
\centering
\begin{minipage}{0.98\textwidth}
\scriptsize
\begin{Verbatim}[
breaklines=true,
breakanywhere=true,
breaksymbolleft={},
breaksymbolright={},
breakindent=0pt,
breakautoindent=false,
fontsize=\scriptsize
]
{
  "id": 30,

  "query": "7. If three points are randomly chosen on a circle, the probability that the triangle formed by these three points is an acute triangle is $\\qquad$ .",

  "sections": [
    {
      "section_id": 1,
      "text": "Okay, so I need to find the probability that three randomly chosen points on a circle form an acute triangle. Hmm, let me recall some geometry here. First, a triangle is acute if all its angles are less than 90 degrees. Now, when dealing with circles,  I remember that an angle inscribed in a circle is related to the arc it subtends. Specifically, the measure of the inscribed angle is half the measure of its subtended arc. So, if all three angles of the triangle are less than 90 degrees, then each of their subtended arcs must be less than 180 degrees because 90 degrees times 2 is 180. Wait, is that right?"
    },
    {
      "section_id": 24,
      "text": "Therefore, the length of y for each x is (x + 1/2) - 1/2 = x. So, the area is the integral from x = 0 to x = 1/2 of x dx. Which is [ (1/2)x^2 ] from 0 to 1/2, which is (1/2)^2 = 1/4.\n\nBut this is the area in the region where x <= y. Since we fixed x <= y, the total area in the entire unit square would be twice this. So, by symmetry, the total area is 2 * 1/4 = 1/2. Therefore, the probability would be 1/2."
    },
    {
      "section_id": 28,
      "text": "But wait, the probability that three random points lie on a semicircle is 3/8. Therefore, the probability that they don't lie on a semicircle is 1 - 3/8 = 5/8. Therefore, the probability that all arcs are less than 1/2 is 5/8. Therefore, the probability of forming an acute triangle is 5/8.\n\nBut I need to verify this. Wait, let me recall that the probability that n points on a circle all lie on some semicircle is n / 2^n. For n=3, that would be 3/8, which matches this result. Therefore, the probability that they don't lie on any semicircle is 1 - 3/8 = 5/8, which would mean that all arcs are less than 1/2. Hence, the probability of forming an acute triangle is 5/8."
    },
    {
      "section_id": 32,
      "text": "Therefore, the probability that three randomly chosen points on a circle form an acute triangle is 1/4.\n\n**Final Answer**\n\\boxed{\\dfrac{1}{4}}"
    }
  ],

  "labels": {
    "errors": [
      {
        "type": "Formal Error",
        "section_start": 24,
        "section_end": 24,
        "section_numbers": [24],
        "description": "Introduces an arithmetic/integration mistake: the integral of x from 0 to 1/2 is incorrectly evaluated as 1/4 instead of 1/8, then used consistently in the symmetry argument."
      },
      {
        "type": "Knowledge Error",
        "section_start": 28,
        "section_end": 28,
        "section_numbers": [28],
        "description": "Misstates the known formula for the probability that n random points lie on a semicircle, using an incorrect denominator and deriving the wrong numerical result."
      }
    ],

    "structure": {
      "layer1": {
        "nodes": [
          {"id": "node1", "type": "problem_decomposition", "section_start": 1, "section_end": 5},
          {"id": "node2", "type": "reasoning_step", "section_start": 6, "section_end": 23},
          {"id": "node3", "type": "intermediate_answer", "section_start": 24, "section_end": 24},
          {"id": "node4", "type": "check", "section_start": 25, "section_end": 25},
          {"id": "node5", "type": "reasoning_step", "section_start": 26, "section_end": 28},
          {"id": "node6", "type": "check", "section_start": 29, "section_end": 30},
          {"id": "node7", "type": "check", "section_start": 31, "section_end": 31},
          {"id": "node8", "type": "final_answer", "section_start": 32, "section_end": 32}
        ],
        "edges": [
          {"from": "node1", "to": "node2", "type": "reasoning"},
          {"from": "node2", "to": "node3", "type": "reasoning"},
          {"from": "node4", "to": "node3", "type": "check"},
          {"from": "node1", "to": "node5", "type": "reasoning"},
          {"from": "node3", "to": "node5", "type": "backtracking"},
          {"from": "node6", "to": "node5", "type": "check"},
          {"from": "node5", "to": "node6", "type": "reasoning"},
          {"from": "node6", "to": "node7", "type": "reasoning"},
          {"from": "node7", "to": "node2", "type": "check"},
          {"from": "node7", "to": "node8", "type": "reasoning"}
        ]
      }
    }
  }
}
\end{Verbatim}
\end{minipage}
\caption{Complete LensBench case annotation for ID 30, including the data schema, original section texts with ASCII-only normalization, detected errors, and structured reasoning graph.}
\label{fig:lensbench_case_id56_full}
\end{figure*}

\end{document}